\RequirePackage{snapshot}

\documentclass[runningheads]{llncs}
\usepackage[utf8]{inputenc}
\usepackage{xcolor}
\usepackage{amsmath}
\usepackage{hyperref}
\usepackage{graphicx}
\usepackage{subcaption}
\usepackage{overpic}
\usepackage{tikz}
\usepackage[normalem]{ulem}

\newcommand{\postPDF}{\ensuremath{F}}

\newcommand{\param}{\ensuremath{\vec{m}}}
\newcommand{\data}{\ensuremath{\vec{d}}}
\newcommand{\cdens}[2]{q\left(#1|#2\right)}
\newcommand{\MCmove}[2]{b\left(#1, #2\right)}
\newcommand{\hastingsR}[2]{r\left(#1,#2\right)}
\newcommand{\lrp}[1]{\left(#1\right)}
\newcommand{\lrb}[1]{\left[#1\right]}
\newcommand{\NormDist}[1]{\ensuremath{\mathcal{N}\lrp{#1}}}

\newcommand{\fwd}[1]{g\lrp{#1}}
\newcommand{\grad}{\ensuremath{G}}
\newcommand{\datanoise}{\ensuremath{\vec{\epsilon}}}
\newcommand{\ensize}{N}
\newcommand{\rw}[1]{{\color{red} #1}}

\begin{document}

\title{Deep learning for prediction of complex geology ahead of drilling}
\author{Kristian Fossum\inst{1}, Sergey Alyaev\inst{1}, Jan Tveranger\inst{1}, Ahmed Elsheikh\inst{2}}
\authorrunning{K. Fossum et al.}

\institute{NORCE Norwegian Research Centre, Bergen, Norway \\
  \and
School of Energy, Geoscience, Infrastructure and Society, Heriot-Watt University, Edinburgh, United Kingdom}

\maketitle

\begin{abstract}
During a geosteering operation the well path is intentionally adjusted in response to the new data acquired while drilling. To achieve consistent high-quality decisions, especially when drilling in complex environments, decision support systems can help cope with high volumes of data and interpretation complexities. They can assimilate the real-time measurements into a probabilistic earth model and use the updated model for decision recommendations.

Recently, machine learning (ML) techniques have enabled a wide range of methods that redistribute computational cost from on-line to off-line calculations.   
In this paper, we introduce two ML techniques into the 
geosteering decision support framework. 
Firstly, a complex earth model representation is generated using a Generative Adversarial Network (GAN). 
Secondly, a commercial extra-deep electromagnetic simulator is represented using a Forward Deep Neural Network (FDNN).


The numerical experiments demonstrate that 
the combination of the GAN and the FDNN in an ensemble randomized maximum likelihood data assimilation scheme provides real-time estimates of complex geological uncertainty.
This yields reduction of geological uncertainty ahead of the drill-bit from the measurements gathered behind and around the well bore.

\end{abstract}

\section{Introduction}
\label{sec:introduction}
The process of drilling wells for hydrocarbon production represents a major cost in petroleum reservoir development. 
However, drilling of new wells is necessary to increase the total oil recovery. 
To maximize the value for each drilled well it is necessary to optimize the placement of the well within the reservoir structure. 
An optimally placed well will mobilize more of the petroleum resources, and reduce the need for injected water \rw{--} reducing the environmental impact of oil production.

To place a well in its optimal position, operators apply geosteering. 
Here, the well trajectory is adjusted while drilling in response to real-time measurement of the geology surrounding the drill bit. 
The value of geosteering has been well documented in the literature~\cite{Al-Fawwaz2004,Guevara2012,Janwadkar2012}.

The main objective with geosteering is to utilize the information in the measurements to make optimal decisions. 
Hence, geosteering can be seen as a sequential decision process under uncertainty and should be treated in a probabilistic framework \cite{kullawan2014decision}.
Recently, a workflow based on the Ensemble Kalman Filter (EnKF)~\cite{Evensen1994} 
has been employed to condition the geological model on measurements acquired while drilling~\cite{Chen2015b,Luo2015}.
In the EnKF the uncertainty is represented
by an ensemble of equiprobable realizations.
This workflow has then been combined with a global optimization method and
applied as a Decision Support System (DSS)\cite{Alyaev2019a}.

The DSS framework  provides high quality decisions on synthetic cases, but practical challenges should be addressed for it to be applicable to real operations~\cite{Alyaev2019a}.
Firstly, to our knowledge, there is no studies which combine the ensemble update with a commercial tool for simulation of measurements. 
Secondly, the earth model utilized in the published studies does not represent \sout{a} realistic geological complexity. 
Conceptually, it is easy to insert any numerical model for simulating the measurements into the DSS workflow. 
Similarly, there is nothing that prohibits the use of complex earth models. 
However, as the complexity increase, the numerical run-time also increases hindering real-time performance.
Moreover, complex earth models can not typically  be represented using a Gaussian distribution, and consequently EnKF updates 
will  not retain the geological complexity \cite{Sebacher2015}.

\begin{figure}
    \centering
    \includegraphics[width=\textwidth]{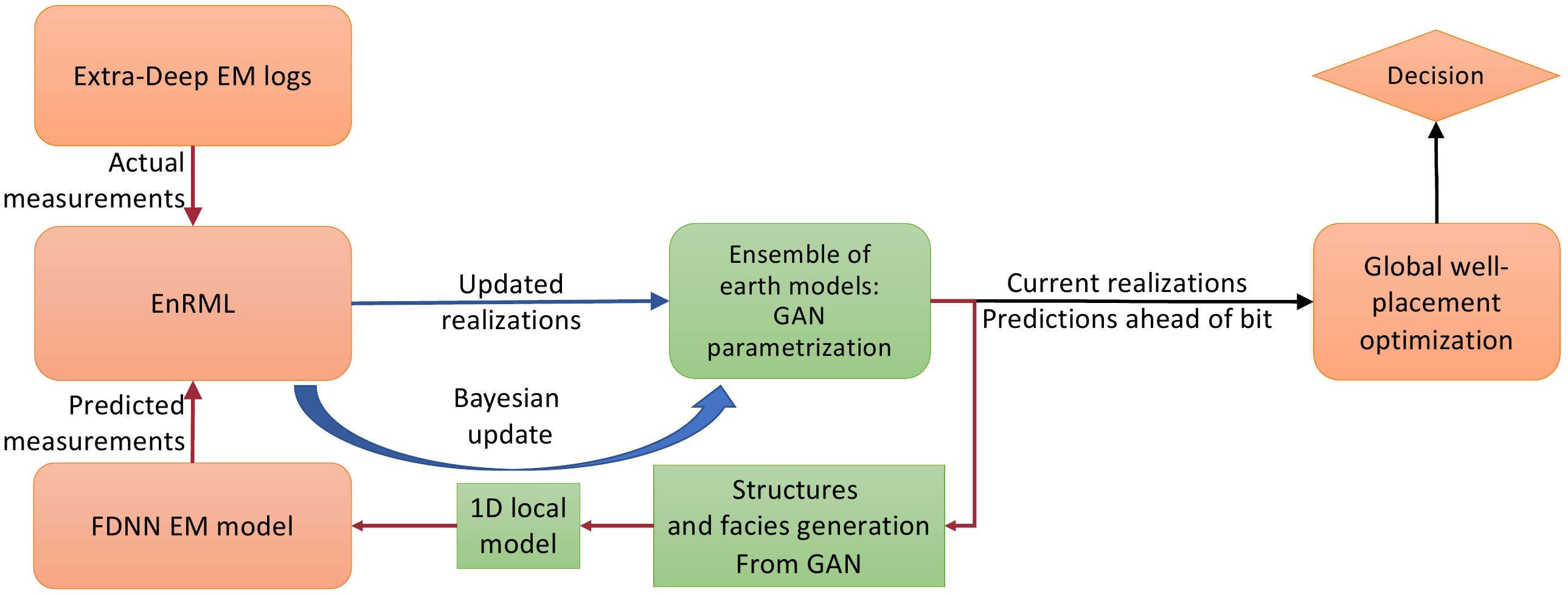}
    \caption{The proposed DSS workflow. Green boxes highlight the new elements introduced in this paper.}
    \label{fig:DSS_workflow}
\end{figure}

In this paper, we introduce important elements to make the DSS better suited to real operations, see Figure~\ref{fig:DSS_workflow}.
The main novelty in our approach is to introduce a machine-learning method to represent both 
the earth model and the forward model of extra-deep borehole electromagnetic (EM) measurements.
Within this setting we demonstrate that a real-time ensemble-based inversion 
can indeed predict the distribution of  non-trivial geology ahead of bit. 

To construct a reference earth model 
we generate realizations of a fluvial geological environment using a commercial software.
These realizations are then sub-sampled to form a training dataset for the offline training of a Generative Adversarial Network (GAN). 
The GAN is then used, online, to generate plausible geological realizations from a low-dimensional Gaussian input vector. The earth modeling is described in Section~\ref{sec:GAN}.
For modeling the extra-deep EM measurements we use a forward deep neural network (FDNN) trained on a dataset generated using a commercial simulator~(Section~\ref{sec:proxyLog}).
%
%
%
In Section~\ref{sec:DA} we discuss the exact and the fast data assimilation (DA) methods. 
The numerical results, showing the applicability of our proposed method are given in Section~\ref{sec:num_inv}. 
Finally, we summarize and conclude the paper in Section~\ref{sec:sum_conc}.

\section{Earth modeling using GAN}
\label{sec:GAN}

GANs are a class of unsupervised machine learning methods which can learn to generate new formatted data with the same statistics as the training set.
Motivated by successful applications of GANs for modeling channelized structures for reservoir simulation \cite{Chan2019,Chan2019a}, we use a GAN for efficient earth modeling.

The GAN  consists of two deep neural networks (DNNs): a generator and a discriminator. 
The generator takes a random Gaussian low dimensional vector as input and generates a realization of formatted data: geological realization.
The discriminator takes the formatted data and gives a probability of it being 'real', i.e., belonging to the training set. 
During training the DNNs contest each other in a min-max game.
They are trained simultaneously. 
On each training step the generator creates (fake) geological realization from random vectors. 
Fake geological realizations are combined with random samples of real earth model and are fed to the discriminator.
The loss function for the  generator is  proportional to number of 'fakes' correctly identified by the discriminator.
The loss function for the discriminator is proportional to the total misjudged data samples. 
In our study we use an adapted Wasserstein GAN with hierarchical convolutional networks for the generator and the discriminator, see \cite{Arjovsky2017} for implementation details. 

\begin{figure}
    \centering
    \includegraphics[width=0.95\textwidth]{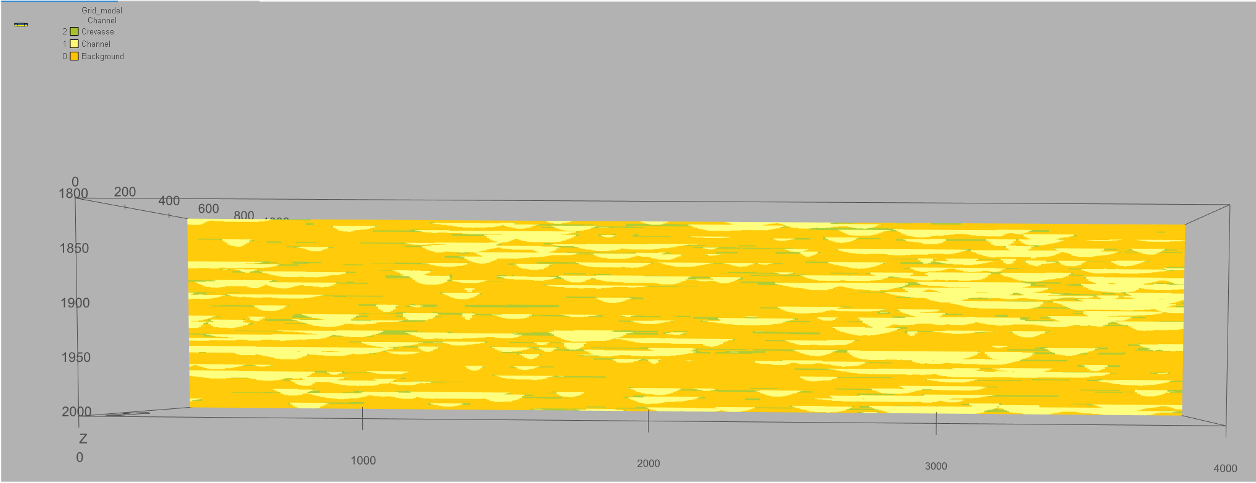}
    \caption{The original earth model generated by the commercial tool.}
    \label{fig:originalModel}
\end{figure}

For geosteering we want to reproduce likely geological realizations of facies and porosity distributions on a 2D vertical geological section along the well to identify the oil-bearing sands ahead of bit.
For training of GAN we use a large (compare to the area of prediction) reference earth model,
which should provide a realistic test case for the present study in terms of scale and actual geological features and properties.
The reference earth model is constructed using a commercial software that models a synthetic structural framework, a facies model set-up derived from outcrop analogue data, and synthetic petrophysical properties of individual facies derived from published literature. 
The resulting model measures 4000m x 1000m x 200m (xyz) with cell
dimensions set to 10x10x0.5~m, yielding a regular  corner-point  grid of size 400x100x400, see Figure~\ref{fig:originalModel}.

The constructed facied model represents a low net/gross fluvial depositional system. 
It was chosen since it provides complex 3D architectures comprising a limited number of facies, which form contrasting 
geometries, see Figure~\ref{fig:originalModel}.  
Input numbers for statistical generation of facies and geometries are derived from a well-documented outcrop of the Cretaceous lower Williams Fork Formation (Mesa Verde Group) at Coal Canyon, Colorado, USA~\cite{Pranter2014,Pranter2011,Trampush2017}.
%

Key parameters of the facies model set-up are listed in Table~\ref{tab:tab1}. The model is not intended as a rendering of the outcrop itself and is consequently simplified compared to descriptions of the original outcrop~\cite{Cole2005,Panjaitan2006,Pranter2009,Pranter2011}.
The model contains three facies: Background/shale, Channels and Crevasse splays. 
The probability distribution of channel width in the model is adapted to include “narrow channel bodies”, and stacking of channels accounts for multi-story channels which comprise more than 80\% of the observed channel bodies. The flow direction of the channel system is set towards $45 \pm 10$ degrees. No trends were used to condition the spatial distribution of channels.  

\begin{table}
\caption{Parameter settings for facies models.}\label{tab:tab1}
\begin{tabular}{|l|l|l|l|}
\hline
\underline{Volumetric fraction} & Value & Tolerance & Comments\\
\hline
Channel system volume fraction &  0.3 & 0.05 & \\
Channel positioning &  1 & & No trends \\
  Crevasse volume fraction & 0.1 & 0.03 & Of channel system vol. frac. \\
  \hline
\end{tabular}
\begin{tabular}{|l|l|l|l|l|}
\hline
\underline{Channel geometry} & Value & SD & Min. & Max.\\
\hline
  Thickness & 4.2 & 1.5 & & \\
  Width & 155 & 50 & 20 & 500 \\
  Correlation W/T & 36 & & & \\
  Amplitude & 400 & 50 & & \\
  Sinuosity & 1.3 & & & \\
  Azimuth & 45 & 10 & & \\
  \hline
\end{tabular}
\begin{tabular}{|l|l|}
\hline
\underline{Form/repulsion} & Setting  \\
\hline
  Cross-section geometry & Parabolic, basic variability \\
  Channel form & Rigid \\
  Repulsion & None \\
  \hline
\end{tabular}
\end{table}

The geological realization is parameterized by a vector of 60 independent parameters.
For each 60-dimensional vector, the generator outputs a 64x64 grid with three values in each grid block.
For a grid block (with dimensions 10.0m along-well and 0.5m thickness) the three values, 'channels',  represent the 
probability of the grid-block belonging to the respective facies class: Background/Channel/Crevasse.
Our generator is also predicting porosity/resistivity distribution within the geo-bodies, but in this initial study only the facies classes are used.

For training, the original 3D earth model is sampled as 64x64 2D images with three channels. 
The  facies index from the training set is converted into one-hot three-dimensional vector. 
That is, the vector represents the probability of facies: the value of the true index is set to one and other channels to zero.
During evaluation, the resistivity of the facies with highest probability is applied.




\section{LWD Neural Network}
\label{sec:proxyLog}
To maintain real-time performance of a data assimilation workflow the forward model should be 
fast and support batch, preferably parallel execution. 
Proprietary forward models provided by measurement instrument vendors provide the most accurate results, but they are often not sufficiently fast, and not always optimized for batch execution.
In \cite{alyaev2020modeling}, the authors developed a DNN approximation of such a forward model~\cite{Sviridov2014a}, which we abbreviate FDNN. 

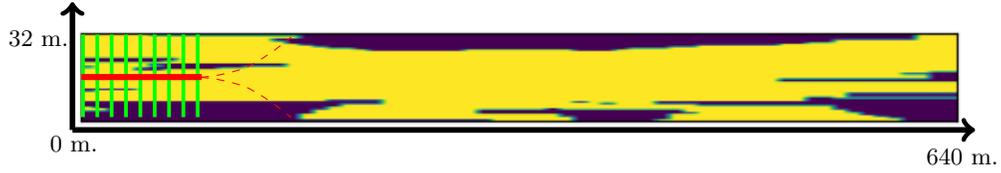
\begin{figure}
    \centering
    \begin{overpic}[width=\textwidth]{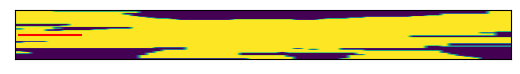}
    \put(0.75,1.5){
    \begin{tikzpicture}
    \draw[line width=0.75mm, ->] (0,0) -- (12,0);
    \draw[line width=0.75mm, ->] (0,0) -- (0,1.7);
    \end{tikzpicture}
    }
    \put(95,-2){640 m.}
    \put(-5,11) {32 m.}
    \put(-.5,-0.5){0 m.}
    \put(2,3){
    \begin{tikzpicture}
    \draw[line width=0.5mm,green] (0,0) -- (0,1.1);
    \end{tikzpicture}}
    \put(3.56,3){
    \begin{tikzpicture}
    \draw[line width=0.5mm,green] (0,0) -- (0,1.1);
    \end{tikzpicture}}
    \put(5.12,3){
    \begin{tikzpicture}
    \draw[line width=0.5mm,green] (0,0) -- (0,1.1);
    \end{tikzpicture}}
    \put(6.68,3){
    \begin{tikzpicture}
    \draw[line width=0.5mm,green] (0,0) -- (0,1.1);
    \end{tikzpicture}}
    \put(8.25,3){
    \begin{tikzpicture}
    \draw[line width=0.5mm,green] (0,0) -- (0,1.1);
    \end{tikzpicture}}
    \put(9.81,3){
    \begin{tikzpicture}
    \draw[line width=0.5mm,green] (0,0) -- (0,1.1);
    \end{tikzpicture}}
    \put(11.375,3){
    \begin{tikzpicture}
    \draw[line width=0.5mm,green] (0,0) -- (0,1.1);
    \end{tikzpicture}}
    \put(12.9375,3){
    \begin{tikzpicture}
    \draw[line width=0.5mm,green] (0,0) -- (0,1.1);
    \end{tikzpicture}}
    \put(14.5,3){
    \begin{tikzpicture}
    \draw[line width=0.5mm,green] (0,0) -- (0,1.1);
    \end{tikzpicture}}
    \put(1.7,7.25){
    \begin{tikzpicture}
    \draw[line width=0.8mm,red] (0,0) -- (1.6,0);
    \end{tikzpicture}}
    \put(15,7.5){
    \begin{tikzpicture}
    \draw[red,dashed] (0,0) .. controls (0.2, 0.01398536238) and (0.4, 0.05621633388) .. (0.6, 0.127533937) .. controls (0.8, 0.2293963086) and (1,0.36397023426) .. (1.2, 0.53427442237); 
    \end{tikzpicture}
    }
    \put(15,3.1){
    \begin{tikzpicture}
    \coordinate (p1) at (0,0);
    \coordinate (p2) at (0.2,-0.013985362380);
    \coordinate (p3) at (0.4,-0.05621633388);
    \coordinate (p4) at (0.6,-0.127533937);
    \coordinate (p5) at (0.8,-0.2293963086);
    \coordinate (p6) at (1, -0.36397023426);
    \coordinate (p7) at (1.2, -0.53427442237);
    \draw[red, dashed] (p1) .. controls (p2) and (p3) .. (p4) .. controls (p5) and (p6) .. (p7);
    \end{tikzpicture}
    }
\end{overpic}
    \caption{Resistivity of earth model plotted in 1:2 aspect ratio. The green lines shows the measurements and their extent illustrate the maximum sensitivity depth. The full red line is the drilled well, and the dashed red lines indicate the potential for geosteering.}
    \label{fig:DrillGAN}
\end{figure}

The model approximates the output of the ultra-deep electromagnetic well-bore logging instrument. The instrument is configured to transmit four shallow and nine pairs of deep directional measurements, and has sensitivity to boundaries up to 30 meters to the side from the well bore. We emphasize that the tool provides information around, but not ahead of the drilling position. An illustration of the deep measurements is provided in Figure~\ref{fig:DrillGAN}.

The input to the FDNN model is a layered geological media with up to three boundaries above and below the measurement instrument as well as the resistivity values of all seven layers. 
In this study we assume that the layer resistivity is isotropic and that the well is aligned with the horizontal axis. 

We produce one synthetic set of measurements for every horizontal position of the gridded model which we 'drill' through. 
First, we choose most probable facies for each 'pixel' and substitute it with the corresponding resistivity value: 
\begin{enumerate}
    \item Background, $R = 220.0$ Ohm m;
    \item Channel, $R = 3.6$ Ohm m;
    \item Crevasse, $R = 4.1$ Ohm m.
\end{enumerate}
Second, we find boundaries between layers composed of pixels with equal resistivities and use them as the input to the forward model.







\section{Data assimilation}
\label{sec:DA}



In the DSS for geosteering \cite{Alyaev2019a}, one uses data assimilation to condition the earth model to measurements made while drilling. The fundamental idea is that if a poorly known earth model can be made consistent with measurements it will provide more accurate forecasts, and, hence, provide a better basis for decisions.

In this paper, the emphasis is placed on the data assimilation part of the DSS. Especially data assimilation utilizing an efficient neural network model for the synthetic logs, and an efficient GAN-generator for representing complex earth models. To check whether this setup provides useful conditioned models we consider two data assimilation algorithms. Firstly, the Markov Chain Monte Carlo (MCMC), which is considered as a gold standard method for sampling. Secondly, the ensemble randomized maximum likelihood (EnRML), an approximate method suitable for DSS.

\subsection{MCMC}

A reliable method for sampling from a complex posterior distribution is the MCMC technique. MCMC relates to the general framework of methods introduced in~\cite{Metropolis1953} and~\cite{Hastings1970a} for Monte Carlo (MC) integration. 
One designs a Markov chain that produce samples from the desired posterior distribution, and subsequently utilize these samples for MC estimation. In this section, the adaptive Metropolis-Hastings method, utilized in the numerical study, is introduced. For more information on MCMC we refer the reader to~\cite{Brooks2011}, and references therein.

Suppose we want samples from the un-normalized posterior distribution \postPDF, which is the general case with the Bayesian method where the normalizing factor often is very difficult to calculate. 
Assume that the current element is \param, and that the chain proposes a move to $\param^*$, with conditional probability density $\cdens{\param^*}{\param}$. The move is performed with probability 
\begin{equation}
\MCmove{\param}{\param^*} = min \left(1, \hastingsR{\param}{\param^*} \right)     
\end{equation}
where the Hastings ratio is defined as
\begin{equation}
    \hastingsR{\param}{\param^*} = \frac{\postPDF\lrp{\param^*} \cdens{\param}{\param^*}}{\postPDF\lrp{\param}\cdens{\param^*}{\param}}.
\end{equation}
If the move is not made $\param^* = \param$. This is the basis for the Metropolis-Hastings method, and it can be shown that the method will generate samples from the posterior distribution $\postPDF$. 

The Metropolis-Hastings algorithm requires a choice of proposal distribution, and some distributions will work better than others. Intuitively, one would like to draw proposal samples from \postPDF. However, this is not possible since we cannot sample from this distribution. However, one idea is to consider the previous samples from the algorithm as approximate samples from \postPDF. With this approach proposal samples are drawn from
\begin{equation}
    \param^* \sim \lrp{1-\beta}\NormDist{\param,\lrp{\frac{2.38^2}{N_m}}\tilde{C}_{\param}} + \beta \NormDist{\param, Q_{\param}},
\end{equation}
where $\tilde{C}_{\param}$ is the empirical covariance matrix calculated utilizing all the preceding iterations of the Markov Chain, $Q_{\param}$ is some fixed non-singular matrix and $0 < \beta < 1$. Note that $\beta = 1$ until $C_{\param}$ is well defined. This sampling method was applied in~\cite{Fossum2014b,Fossum2015}.
It is well known that the MCMC requires a certain burn-in period, since the initial samples are not from the posterior distribution. Hence, it is necessary to monitor the convergence of the method. In this work, convergence is monitored by assessing the maximum root statistic of the multivariate potential scale reduction factor~\cite{Brooks1998}.


\subsection{EnRML}
\label{sec:enrml}
The EnRML~\cite{Gu2007} has recently become one of the most successful methods for automatic history matching of petroleum reservoirs. The EnRML is based on minimization of an objective function using the ensemble approximation of the sensitivity matrix. Hence, the EnRML can be formulated in many different ways. In this study we utilize  the approximate form of the Levenberg-Marquardt method, introduced in~\cite{Chen2013}.


Iteration number $i$ of the Levenberg-Marquardt method is given as
\begin{align}
    \delta \param_i =& - \lrb{\lrp{1 + \lambda_i}C^{-1}_{\param} + \grad^T_i C^{-1}_{\data} \grad_i}^{-1} \\
            & \times \lrb{C^{-1}_{\param}\lrp{\param_i - \param_{prior}} + \grad^T_i C_{\data}^{-1} \lrp{\fwd{\param} - \lrp{\data_{obs} + \datanoise}}}
\label{eq:LM_full_update}
\end{align}
where $\lambda_i$ is the Levenberg-Marquardt multiplier, $\grad$ is the sensitivity of data to the parameters, and $\datanoise\sim\NormDist{0,C_{\data}}$ is a realization of the measurement observation noise. 

In the ensemble framework, we approximate $C_{\param}$ and $\grad$ using the ensemble. To this end we define 
\begin{equation}
    \tilde{\grad} = C^{1/2}_{sc}\Delta \data \lrp{\Delta \param}^{-1}
\end{equation}
\begin{equation}
    \tilde{C}_{\param} = \Delta \param \Delta \param^T
\end{equation}
where 
\begin{equation}
    \Delta \param = \lrb{\param_1, \dots, \param_j, \dots, \param_{\ensize}}\lrp{I_{\ensize} - \frac{1}{\ensize}11^T}/\sqrt{\ensize - 1},
\end{equation}
\begin{equation}
    \Delta \data = C_{sc}^{-1/2}\lrb{\fwd{\param_1}, \dots, \fwd{\param_j}, \dots, \fwd{\param_{\ensize}}}\lrp{I_{\ensize} - \frac{1}{\ensize}11^T}/\sqrt{\ensize - 1},
  \end{equation}
$\ensize$ denotes the ensemble size, and $C_{sc}$ is a diagonal matrix for scaling the data, typically containing the measurement variance on the diagonal.
We get the approximate version of the Levenberg-Marquardt update equation by inserting ensemble approximations of $\grad$ and $C_{\param}$,  neglecting the updates from the model mismatch term, substituting the prior precision matrix $C^{-1}_m$ with $\tilde{C}^{-1}_{\param_i}$, and rewriting the equation using the Sherman-Woodbury-Morrison matrix inversion formula~\cite{Golub1983}
\begin{equation}
  \label{eq:LM_approx_update}
    \delta \param_i = - \tilde{C}_{\param_i}\tilde{\grad}_i^T\lrb{\lrp{1+\lambda_i}C_{\data} +  \tilde{\grad_i}\tilde{C}_{\param_i}\tilde{\grad}^T_i}^{-1}\lrp{\fwd{\param} - \lrp{\data_{obs} + \datanoise}}.
\end{equation}
The update equation is simplified by calculating the truncated singular value decomposition of $\Delta d$
\begin{equation}
  \Delta \data = U_p S_p V_p^T,
\end{equation}
where the subscript $p$ indicates the number of singular values that are kept. In this work, we define $p$ such that the cumulative sum of the $p$ singular values equals 99\% of the cumulative sum of all the singular values. Further, to allow for correlated measurement errors, we substitute $C_D$ with the ensemble approximation $\tilde{C}_D$
\begin{equation}
  \tilde{C}_{\data}=\Delta \datanoise \Delta \datanoise^T,
\end{equation}
where
\begin{equation}
  \Delta \datanoise = \lrb{\datanoise_1, \dots, \datanoise_j, \dots, \datanoise_{\ensize}}\lrp{I_{\ensize} - \frac{1}{\ensize}11^T}/\sqrt{\ensize - 1}.
\end{equation}
Inserted into~\eqref{eq:LM_approx_update} gives
\begin{align}
  \label{eq:LM_approx_final_one}
  \delta \param_i = -& \Delta \param_i V_p \lrb{\lrp{1 + \lambda_i}S_p^{-1}U_p^TC^{-1/2}_{scl}\Delta \datanoise \Delta \datanoise^TC_{scl}^{-1/2}U_pS_P^{-T} + I} \nonumber \\
                     &\lrp{U_pS^{-1}_p}^TC^{-1/2}_{sc}\lrp{\fwd{\param} - \lrp{\data_{obs} + \datanoise}} \\
                  = -& \Delta \param_i V_p Z \lrb{\lrp{1 + \lambda_i}\zeta + I}^{-1}\lrp{U_pS^{-1}_p Z}^T C^{-1/2}_{sc}\lrp{\fwd{\param} - \lrp{\data_{obs} + \datanoise}},  \nonumber
\end{align}
where $Z$ and $\zeta$ are the eigenvectors and eigenvalues of $S_p^{-1}U_p^TC^{-1/2}_{scl}\Delta \datanoise \Delta \datanoise^TC_{scl}^{-1/2}U_pS_P^{-T}$.

The iterative scheme is run until it is converged. Here, we consider the method to be converged when the relative difference in the data misfit is below a given threshold.

\section{Numerical Results}
\label{sec:num_inv}
We, throughout, utilize the generative neural network, introduced in section~\ref{sec:GAN}, to represent the poorly known earth model. Hence, realizations of the  earth model is generated by applying the generative network to parameters sampled from the multivariate standard distribution, $\param\sim\NormDist{0,C_{\param}}$. 
The numerical investigation considers a well drilled horizontally, approximately, in the center of the model and through the 9 first grid-cells. For each drilled grid-cell we simulate measurements using the model introduced in section~\ref{sec:proxyLog}.

\begin{figure}
  \begin{subfigure}[b]{0.24\textwidth}
    \centering
    \includegraphics[width=\textwidth]{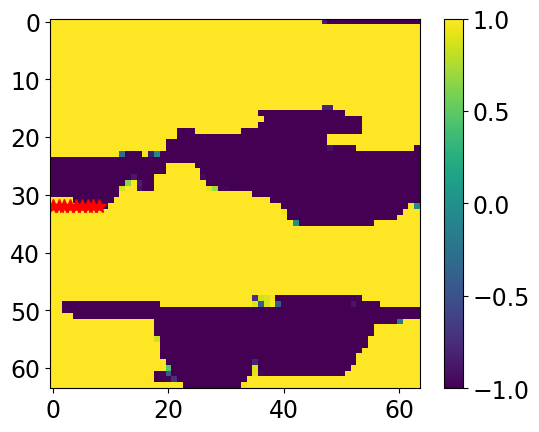}
    \caption{}
  \end{subfigure}
  \begin{subfigure}[b]{0.24\textwidth}
    \centering
    \includegraphics[width=\textwidth]{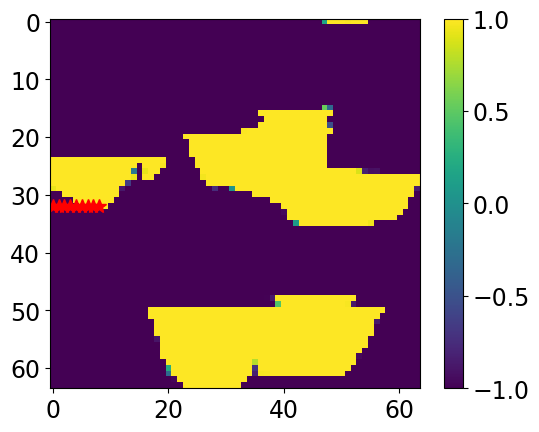}
    \caption{}
  \end{subfigure}
  \begin{subfigure}[b]{0.24\textwidth}
    \centering
    \includegraphics[width=\textwidth]{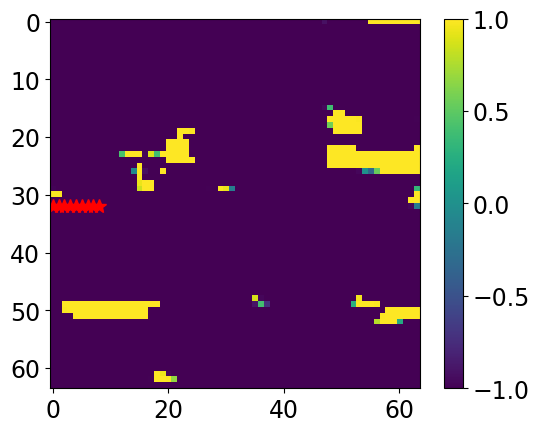}
    \caption{}
  \end{subfigure}
  \begin{subfigure}[b]{0.24\textwidth}
    \centering
    \includegraphics[width=\textwidth]{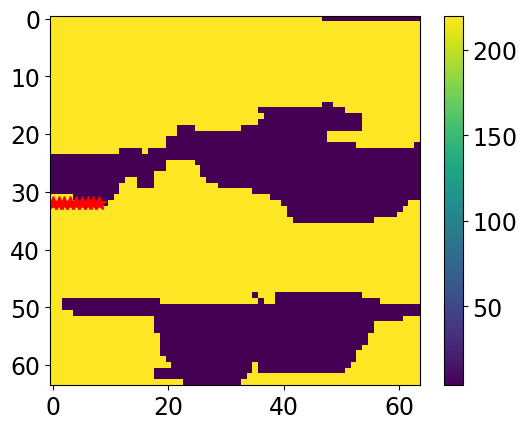}
    \caption{}
  \end{subfigure}
  \\
  \begin{subfigure}[b]{0.24\textwidth}
    \centering
    \includegraphics[width=\textwidth]{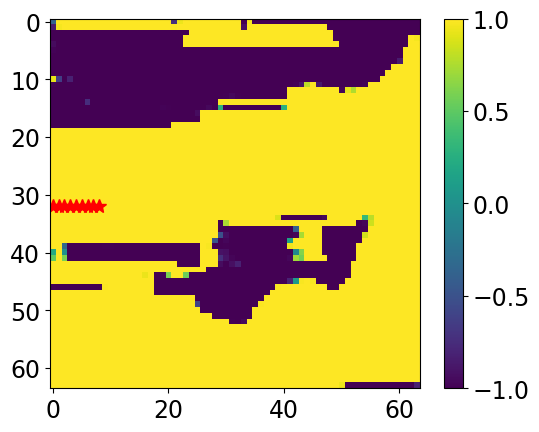}
    \caption{}
  \end{subfigure}
  \begin{subfigure}[b]{0.24\textwidth}
    \centering
    \includegraphics[width=\textwidth]{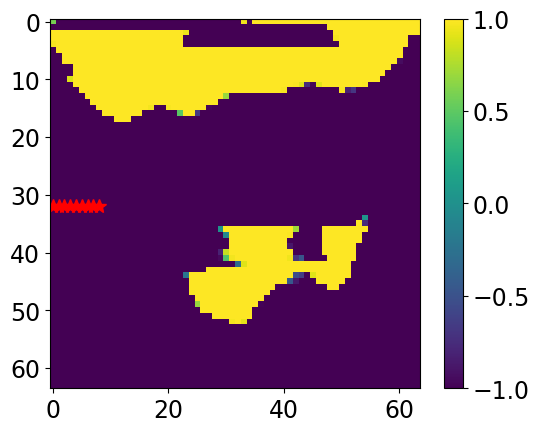}
    \caption{}
  \end{subfigure}
  \begin{subfigure}[b]{0.24\textwidth}
    \centering
    \includegraphics[width=\textwidth]{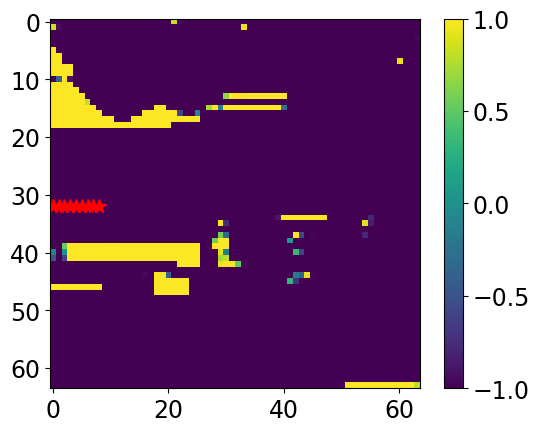}
    \caption{}
  \end{subfigure}
  \begin{subfigure}[b]{0.24\textwidth}
    \centering
    \includegraphics[width=\textwidth]{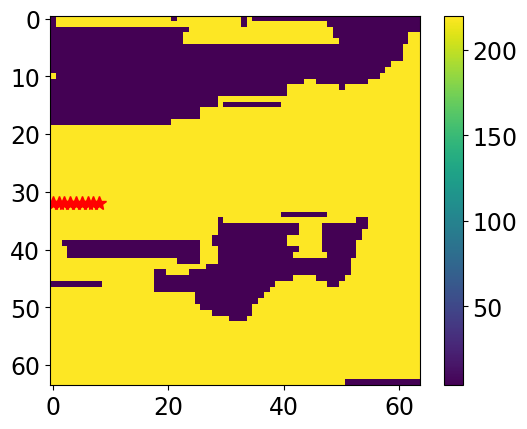}
    \caption{}
  \end{subfigure}
  \\
  \begin{subfigure}[b]{0.24\textwidth}
    \centering
    \includegraphics[width=\textwidth]{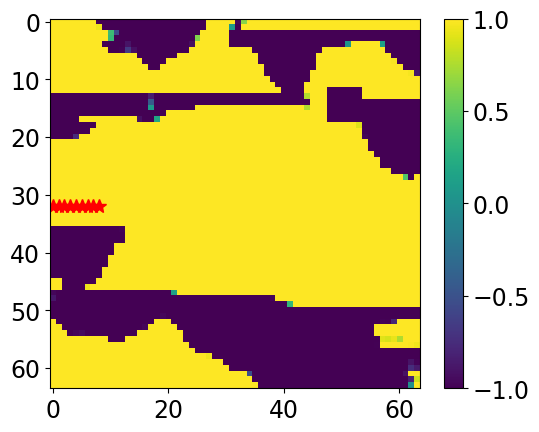}
    \caption{}
  \end{subfigure}
  \begin{subfigure}[b]{0.24\textwidth}
    \centering
    \includegraphics[width=\textwidth]{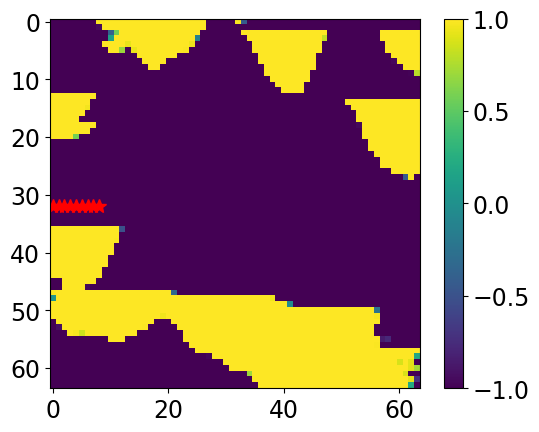}
    \caption{}
  \end{subfigure}
  \begin{subfigure}[b]{0.24\textwidth}
    \centering
    \includegraphics[width=\textwidth]{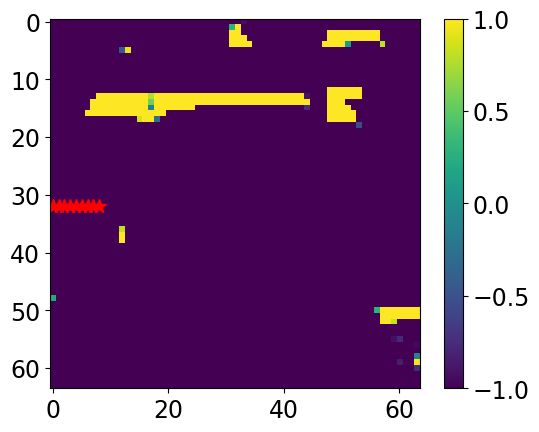}
    \caption{}
  \end{subfigure}
  \begin{subfigure}[b]{0.24\textwidth}
    \centering
    \includegraphics[width=\textwidth]{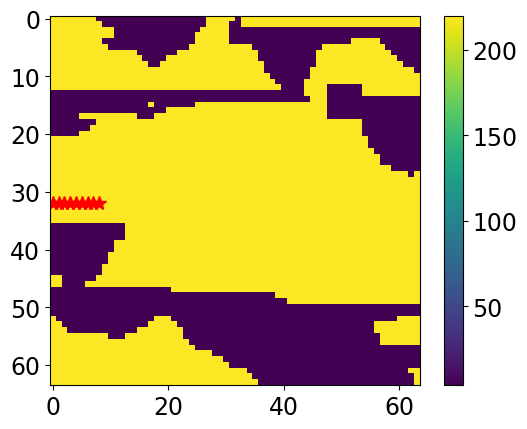}
    \caption{}
  \end{subfigure}
  \caption{Rows 1-3 show three realizations from the prior ensemble.
  Each row shows the three output channels (e.g. fig. a-c) of GAN corresponding to probability of facies (Background/Channel/Crevasse), and the derived resistivity image (e.g. fig. d).
  Red stars indicate measurement position.}
  \label{fig:Prior_models}
\end{figure}

Throughout the investigation, we consider a diagonal $C_{\param}$ with elements equal to $1\times 10^{-6}$. Figure~\ref{fig:Prior_models} shows two random earth model realizations from the prior model. From the figure, we observe that this setup provides significant variation in the earth model. 

\begin{figure}
  \centering
  \begin{subfigure}[b]{0.24\textwidth}
    \centering
    \includegraphics[width=\textwidth]{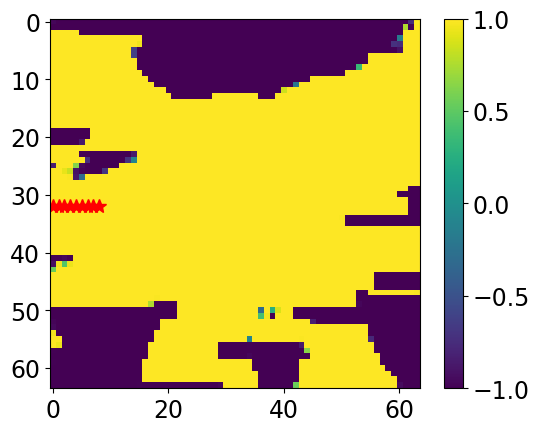}
    \caption{}
  \end{subfigure}
  \begin{subfigure}[b]{0.24\textwidth}
    \centering
    \includegraphics[width=\textwidth]{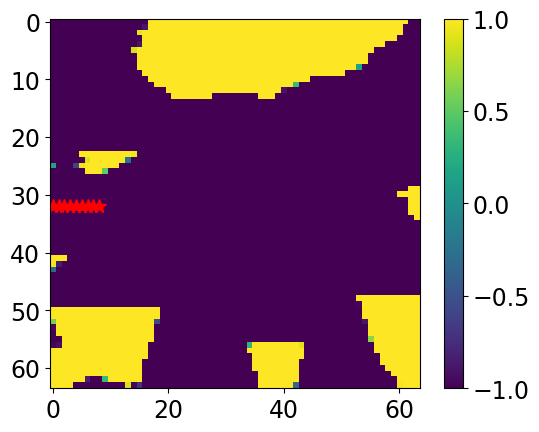}
    \caption{}
  \end{subfigure}
  \begin{subfigure}[b]{0.24\textwidth}
    \centering
    \includegraphics[width=\textwidth]{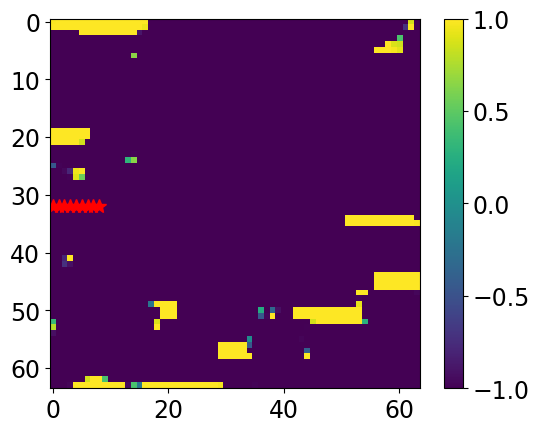}
    \caption{}
  \end{subfigure}
  \begin{subfigure}[b]{0.24\textwidth}
    \centering
    \includegraphics[width=\textwidth]{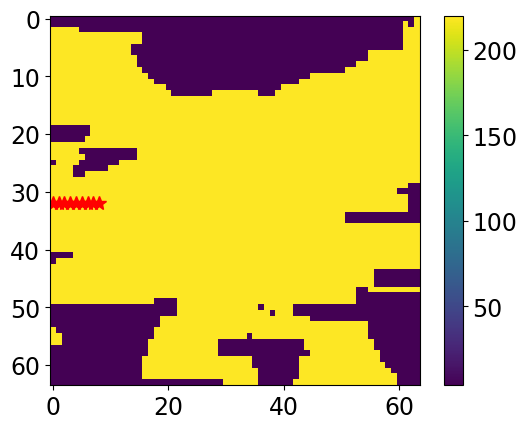}
    \caption{}
  \end{subfigure}
  \caption[True model]{(a)-(c): Facies probabilities from GAN for the synthetic truth. (d): Derived resistivity for the synthetic truth. Red stars indicate measurement positions.}
  \label{fig:True_model}
\end{figure}

The synthetic true earth model is also drawn from the prior model. All 3 channels and the derived resistivity of the synthetic truth is illustrated in figure~\ref{fig:True_model}. The true observations are simulated using the true earth model.

For each of the 9 measurements positions the measurement standard deviation is given as 5\% of the measurement value. In addition, we let the measurements at each position be correlated. Assuming equidistance between measurements, the correlation length equals 10 times the inter-measurement distance. We further assume that the measurements at two different well positions are uncorrelated.

We conduct two numerical experiments. Firstly, we conduct a MCMC run to properly characterize the posterior distribution. Here, 8 chains are run in parallel for $10^{6}$ iteration. At that point the chains were converged, and we extract samples from the posterior by, for each of the 8 chains, removing the first half of the chain and retaining every 100 iteration from the second half of the chain. Hence, leaving $4\times 10^4$ samples from the posterior distribution. 
Secondly, we estimate the posterior distribution using the EnRML method introduced in section~\ref{sec:enrml}. Due to the fast simulation time we utilized an ensemble size of $\ensize=500$, and in addition we applied the correlation based localization technique introduced in~\cite{luo2018b}. The method is allowed to iterate until the relative improvement of the updates is less than $1\%$. When showing the numerical results we will only plot the values of the derived resistivity.
 
\subsection{MCMC}

\begin{figure}
  \centering
  \begin{subfigure}[b]{0.24\textwidth}
    \centering
    \includegraphics[width=\textwidth]{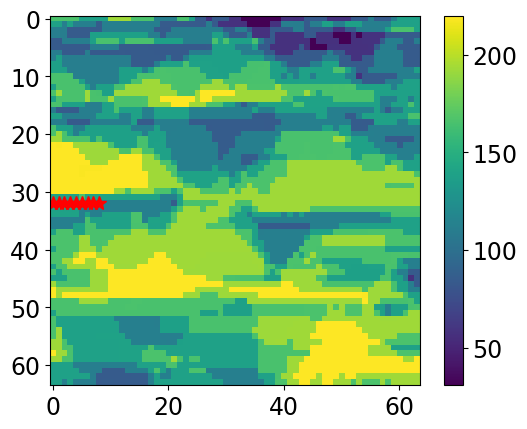}
    \caption{}
  \end{subfigure}
  \begin{subfigure}[b]{0.24\textwidth}
    \centering
    \includegraphics[width=\textwidth]{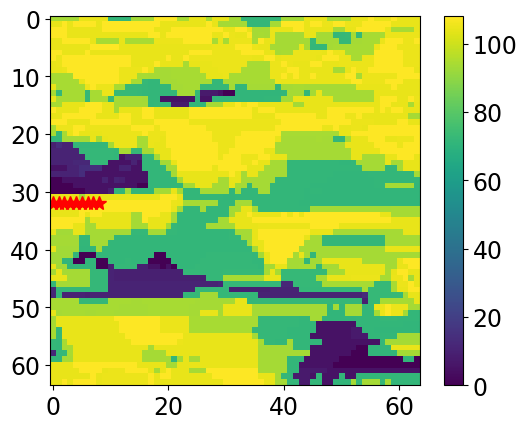}
    \caption{}
  \end{subfigure}
  \begin{subfigure}[b]{0.24\textwidth}
    \centering
    \includegraphics[width=\textwidth]{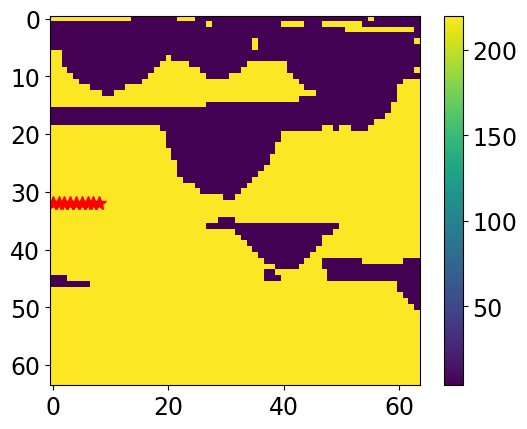}
    \caption{}
  \end{subfigure}
  \begin{subfigure}[b]{0.24\textwidth}
    \centering
    \includegraphics[width=\textwidth]{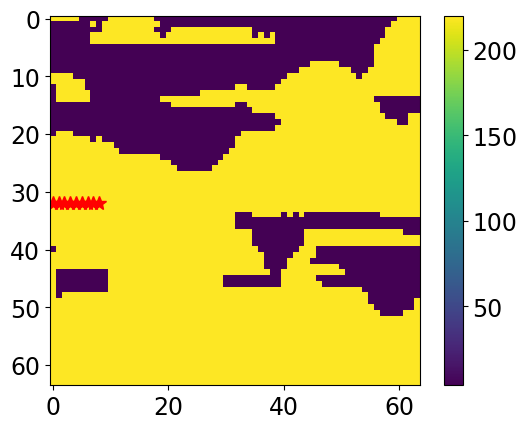}
    \caption{}
  \end{subfigure}
  \caption[MCMC mean and std.]{(a): MCMC mean of the resistivity. (b): MCMC standard deviation of the resistivity. (c): Resistivity for MCMC posterior realization 1. (d): Resistivity for MCMC posterior realization 2.  Red stars indicate measurement positions.}
  \label{fig:MCMC_mean_std and real}
\end{figure}

Based on the posterior realizations obtained by the MCMC, we calculate the posterior mean and posterior standard deviation, shown in figure~\ref{fig:MCMC_mean_std and real} (a) and (b). In figure~\ref{fig:MCMC_mean_std and real} (c) and (d) we plot two random realizations from the posterior.
%

The MCMC result illustrates that the generative neural network can be utilized for data assimilation, and be successfully conditioned to measurements utilizing a neural network proxy model. The model standard deviation is significantly reduced close to the drill bit, and also ahead of the drill bit position. Moreover, the mean value shows that the correct resitivity is identified in these areas. Note that the posterior still has significant variance in most parts of the field. The areas directly around the drill bit have not obtained sufficient reduction of the standard deviation. We do not properly understand why this is so, however, it indicates that there measurements are less sensitive to the region near the drill bit.

\subsection{EnRML}
\begin{figure}
  \centering
  \begin{subfigure}[b]{0.24\textwidth}
    \centering
    \includegraphics[width=\textwidth]{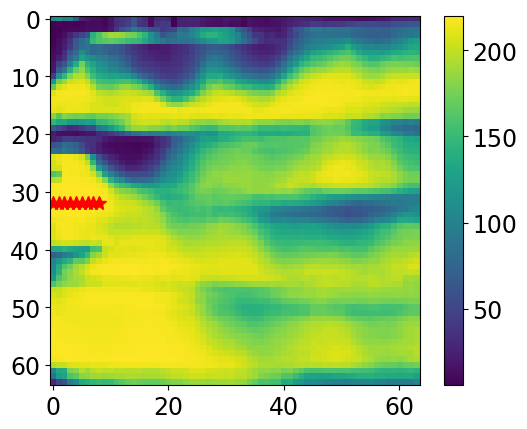}
    \caption{}
  \end{subfigure}
  \begin{subfigure}[b]{0.24\textwidth}
    \centering
    \includegraphics[width=\textwidth]{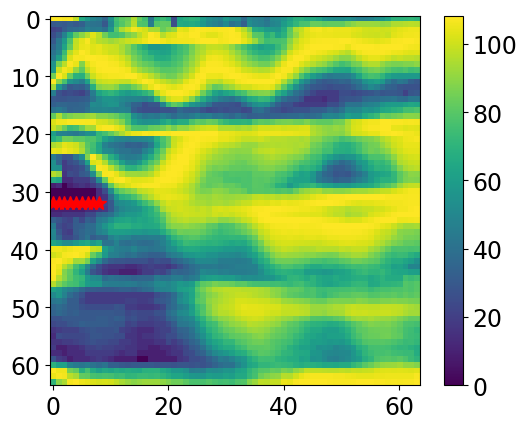}
    \caption{}
  \end{subfigure}
  \begin{subfigure}[b]{0.24\textwidth}
    \centering
    \includegraphics[width=\textwidth]{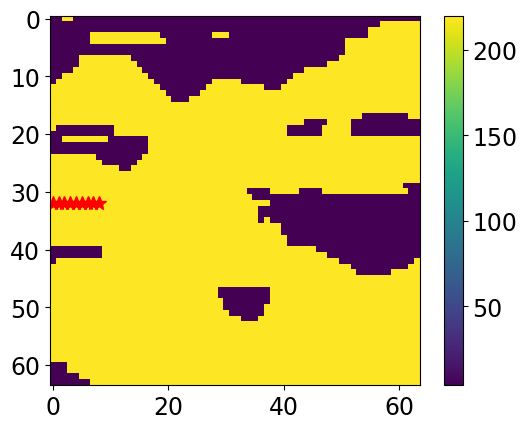}
    \caption{}
  \end{subfigure}
  \begin{subfigure}[b]{0.24\textwidth}
    \centering
    \includegraphics[width=\textwidth]{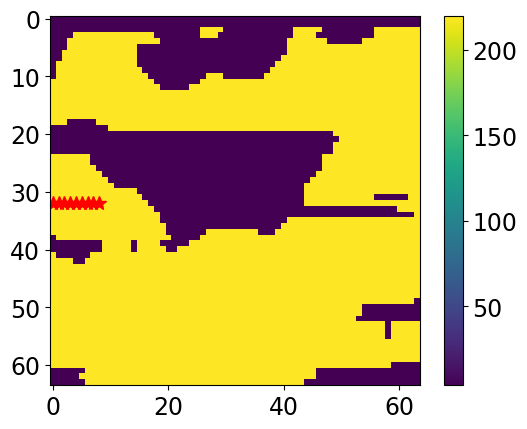}
    \caption{}
  \end{subfigure}
  \caption[EnRML mean and std]{(a): Mean resistivity from the EnRML. (b): Standard deviation of the resistivity from the EnRML. (c): Resistivity for EnRML realization 1. (d): Resitivity for EnRML realization 2. Red stars indicate measurement positions.}
  \label{fig:enrml mean and std real}
\end{figure}

After the EnRML has converged we calculate the mean and standard deviation from the ensemble. These are approximations to the true posterior mean and standard deviations, and are shown in figure~\ref{fig:enrml mean and std real} (a) and (b). In addition, the results from two random realizations are illustrated in  figure~\ref{fig:enrml mean and std real} (c) and (d). 


The results for the EnRML show a significant reduction in the standard deviation around the drill bit, and from the mean model we observe that the correct facies is identified for these regions. There is some reduction of the standard deviation ahead of the drill bit, but we observe that the variance is retained in most of the field.

\section{Summary and conclusions}
\label{sec:sum_conc}
In this paper, we have demonstrated that two essential parts, the earth model and the simulated log, of an ensemble based DSS system can be substituted with neural networks. For the earth model we utilize the GAN trained with images from a realistic geological setting, for the simulated log we use a deep neural network trained using a large set of simulations from a commercial tool. The setup redistributes the computational cost from on-line to off-line calculations, enabling complex earth models utilizing simulated logs with high accuracy to be used in the real time DSS.  
The numerical results illustrate that 
DSS, equipped with the GAN, provides good predictions ahead of drilling when conditioning to only  measurements with sideways sensitivity.

The proposed approach has many beneficial factors. 
Firstly, a GAN provides large flexibility for defining the geological setting. Here, we consider three different facies, but one can easily imagine the inclusion of features like faults and pinch-outs as well as smoothly-varying properties. 
Secondly, we only need to condition a
few parameters with Gaussian distribution
to the measurements,
which is very beneficial for the ensemble based DA approach. 
Thirdly, since we are utilizing a neural network model to generate the simulated log the computational cost of simulating a single ensemble member is very low. Hence, the proposed approach can utilize a large ensemble for the DA part.

The numerical experiments illustrated that the setup provides a reasonable posterior distribution, and we can estimate this using the EnRML approximation. In this study we have only considered a single part of the DSS, namely the conditioning of the earth model to measured data. However, due to the promising results, the developments shown in this paper will integrate with the framework developed in~\cite{Alyaev2019a}. Hence, allowing DSS under much more complex geological setting. Demonstrating the complete DSS with the proposed setup is left for future work.

\section*{Acknowledgments}
The authors are supported by the research project 'Geosteering for IOR' (NFR-Petromaks2 project no. 268122) which is funded by the Research Council of Norway, Aker BP, Equinor, Vår Energi and Baker Hughes Norway.

\bibliographystyle{splncs04}
\bibliography{Geosteering_GAN_Kristian}


\end{document}